\documentclass{article}

\usepackage[preprint]{neurips_2025}

\usepackage[utf8]{inputenc} 
\usepackage[T1]{fontenc}    
\usepackage{hyperref}       
\usepackage{url}            
\usepackage{booktabs}       
\usepackage{amsfonts}       
\usepackage{nicefrac}       
\usepackage{microtype}      
\usepackage{xcolor}         
\usepackage{wrapfig}
\usepackage{graphicx}
\usepackage{amsmath}
\usepackage{multirow}
\usepackage{caption}
\usepackage{underscore}
\usepackage{enumitem}

\title{Efficient Reasoning for LLMs through Speculative Chain-of-Thought}

\author{%
  Jikai Wang \\
  Soochow University \\
  {\small \texttt{risus254@gmail.com}} \\
  \hspace{4.3cm}
  \And
  Juntao Li\thanks{Corresponding author.} \\
  Soochow University,  \\
  {\small \texttt{ljt@suda.edu.cn}} \\
  \hspace{3.5cm}
  \And
  Jianye Hou \\
  CUHK,  \\
  {\small \texttt{jianyehou@link.cuhk.edu.cn}} \\
  \AND
  Bowen Yan \\
  Meta Stone \\
  {\small \texttt{yanbw@meta-stone.com}} \\
  \And
  Lijun Wu \\
  Shanghai AI Laboratory \\
  {\small \texttt{wulijun@pjlab.org.cn}} \\
  \And
  Min Zhang \\
  Soochow University \\
  {\small \texttt{minzhang@suda.edu.cn}} \\
}

\begin{document}

\maketitle

\begin{abstract}
Large reasoning language models such as OpenAI-o1 and Deepseek-R1 have recently attracted widespread attention due to their impressive task-solving abilities.
However, the enormous model size and the generation of lengthy thought chains introduce significant reasoning costs and response latency.
Existing methods for efficient reasoning mainly focus on reducing the number of model parameters or shortening the chain-of-thought length.
In this paper, we introduce Speculative Chain-of-Thought (SCoT), which reduces reasoning latency from another perspective by accelerated average reasoning speed through large and small model collaboration.
SCoT conducts thought-level drafting using a lightweight draft model. Then it selects the best CoT draft and corrects the error cases with the target model.
The proposed thinking behavior alignment improves the efficiency of drafting and the draft selection strategy maintains the prediction accuracy of the target model for complex tasks.
Experimental results on GSM8K, MATH, GaoKao, CollegeMath and Olympiad datasets show that SCoT reduces reasoning latency by 48\%$\sim$66\% and 21\%$\sim$49\% for Deepseek-R1-Distill-Qwen-32B and Deepseek-R1-Distill-Llama-70B while achieving near-target-model-level performance.
Our code is available at \href{https://github.com/Jikai0Wang/Speculative_CoT}{https://github.com/Jikai0Wang/Speculative_CoT}.
\end{abstract}

\section{Introduction}

Large-scale reasoning language models (e.g., OpenAI-o1 \citep{jaech2024openai} and Deepseek-R1 \citep{deepseekai2025deepseekr1incentivizingreasoningcapability}) have exhibited exceptional performance across a broad range of reasoning scenarios.
Their superior performance stems from two key factors:  massive model scale (e.g., DeepSeek-R1 with 671B parameters) and long chain-of-thought (CoT) \citep{wei2023chainofthoughtpromptingelicitsreasoning} generation (reaching over 10,000 tokens for complex tasks).
However, both factors incur substantial computational overhead, resulting in huge inference costs. In many reasoning systems where the model's reasoning process remains implicit, excessive reasoning time induces significant response latency.

Recent works \citep{paliotta2025thinkingslowfastscaling,liao-etal-2025-skintern,xu2025softcotsoftchainofthoughtefficient,zhang2025lightthinkerthinkingstepbystepcompression} mainly tackle this challenge by developing smaller reasoning models or reducing the generation length of CoTs.
Distillation \citep{paliotta2025thinkingslowfastscaling,liao-etal-2025-skintern} serves as a commonly used technique for transferring strong reasoning capabilities from large reasoning models to small ones. Representatively, Deepseek-R1-Distill-Qwen-1.5B \citep{deepseekai2025deepseekr1incentivizingreasoningcapability} exhibits robust performance on diverse datasets.
Approaches such as StepSkip \citep{liucan} and TokenSkip \citep{xia2025tokenskipcontrollablechainofthoughtcompression} suggest training reasoning models on compressed CoT data to shorten the CoT length during inference.
In practical applications, reasoning models need to solve problems with widely varying complexity levels.
Fundamentally, employing large reasoning models to handle simple tasks is a waste.
On the other hand, using small models alone or compressing the chain-of-thought length may compromise performance on complex tasks.
Now that we have a powerful and efficient small model, why not prioritize this lightweight alternative and only invoke the large model for deeper thinking when the problem exceeds the capability threshold of the small one?

\begin{figure}[t]
    \centering
    \includegraphics[width=\textwidth]{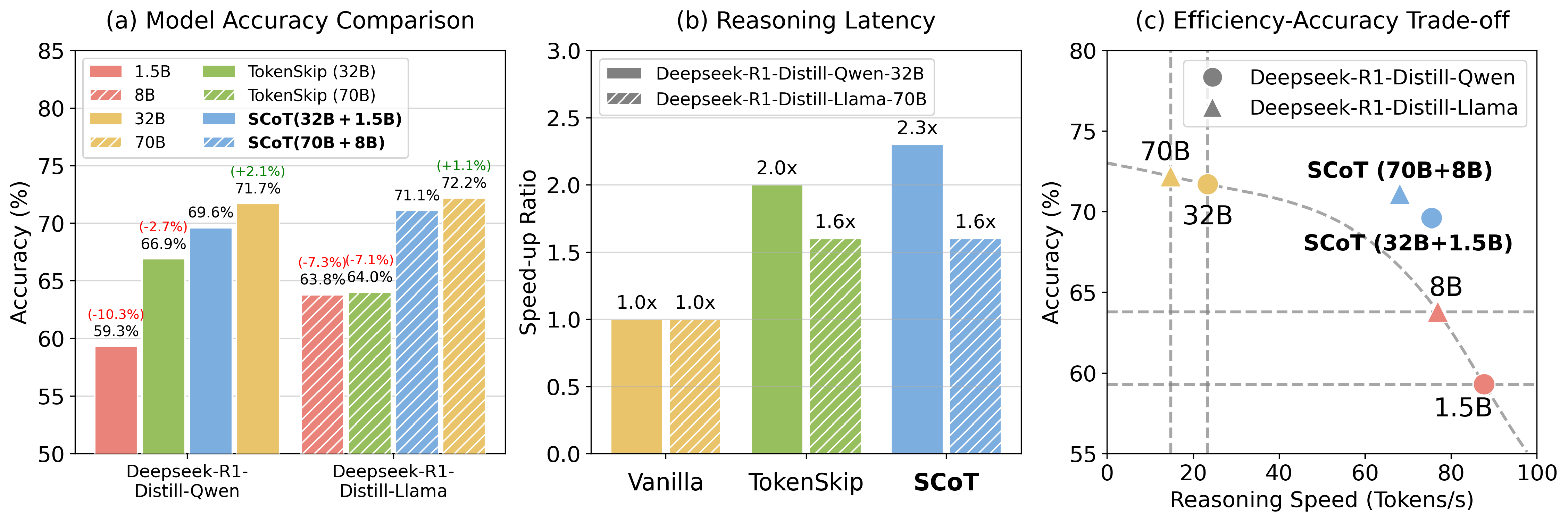}
    \caption{\label{fig:intro} Average reasoning accuracy and efficiency of Speculative Chain-of-Thought (SCoT) with Deepseek-R1-Distill-Qwen-32B and Deepseek-R1-Distill-Llama-B on GSM8K, MATH, GaoKao, CollegeMath and Olympiad. Graph (a): SCoT achieves near-target-model performance level and outperforms TokenSkip on average generation accuracy on the 5 datasets. Graph (b): SCoT demonstrates an average speed-up ratio of 2.3× and 1.6× for the two target models on reasoning latency. Graph (c): Comparison of reasoning speed and accuracy for models of different sizes. Note that all the experiments for testing the reasoning speed are conducted on H800-80G GPUs. SCoT exhibits a better reasoning efficiency-accuracy trade-off through large and small model collaboration. }
\end{figure}

Based on this motivation, we propose Speculative Chain-of-Thought (SCoT), a reasoning framework that reduces reasoning latency through the collaboration of large and small models.
The core idea of SCoT is to use a lightweight model to efficiently generate multiple CoT drafts which are then picked and used by the target model.
SCoT is inspired by speculative decoding \citep{stern2018blockwise,xia2023speculative,leviathan2023fast}, which performs token-level speculation, whereas our approach conducts thought-level speculation.
Unlike shortening the length of CoT, SCoT accelerates the thinking process from another dimension, that is, by increasing the speed of CoT generation.

In Section \ref{sec:scot}, we introduce thinking behavior alignment to improve drafting efficiency.
We also design an error correction mechanism to preserve accuracy in dealing with complex tasks.
We construct a training set by 500 samples from GSM8K to efficiently fine-tuning the target model to improve the capability of draft selection and error detection.
We conduct comprehensive experiments in Section \ref{sec:ex} to evaluate the effectiveness of SCoT.
Figure \ref{fig:intro} displays the overview reasoning accuracy and efficiency of SCoT. 
SCoT reduces reasoning latency by 48\%$\sim$66\% and 21\%$\sim$49\% for Deepseek-R1-Distill-Qwen-32B and Deepseek-R1-Distill-Llama-70B on 5 datasets (including the Olympiad-level dataset \citep{he-etal-2024-olympiadbench}) while maintaining near-target-model-level performance.
It achieves better reasoning accuracy compared with TokenSkip \citep{xia2025tokenskipcontrollablechainofthoughtcompression} under similar speed-up ratios.
Adapting to task difficulty and draft quality, the draft selection mechanism decides when the target model should rethink. Simple tasks mainly utilize the draft model for thought chain generation, while complex problems engage the target model more extensively in the reasoning process.
On the other hand, the single-pass drafting and selection paradigm of SCoT significantly reduces the model reloading overhead between high-bandwidth memory and on-chip cache inherent in traditional speculative decoding methods, maintaining high efficiency even in long-sequence generation scenarios.


\section{Related Works}

\textbf{Efficient Reasoning}\quad
Reasoning LLMs \citep{jaech2024openai,deepseekai2025deepseekr1incentivizingreasoningcapability} generate lengthy CoTs to enhance performance, albeit at the cost of increased latency.
Some approaches \citep{ding2024breakchainlargelanguage,luo2025o1prunerlengthharmonizingfinetuningo1like,xia2025tokenskipcontrollablechainofthoughtcompression,xu2025softcotsoftchainofthoughtefficient,zhang2025lightthinkerthinkingstepbystepcompression} speed up the thinking process by reducing the length of the CoT at the token level. \citet{liucan} and \citet{xia2025tokenskipcontrollablechainofthoughtcompression} use compressed CoTs to fine-tune the reasoning LMs.
\citet{hou2025thinkprunepruninglongchainofthought} prunes long CoTs through reinforcement learning.
\citet{xu2025chaindraftthinkingfaster} and \citet{aytes2025sketch} guide the model to think efficiently through prompting.
Soft CoT methods \citep{saunshi2025reasoning,xu2025softcotsoftchainofthoughtefficient,hao2024traininglargelanguagemodels} use latent space representations to replace reasoning tokens, thus reducing the excessive redundancy in the reasoning process.
Another mainstream approach \citep{yu2024distilling21,paliotta2025thinkingslowfastscaling,liao-etal-2025-skintern} for efficient reasoning is to teach smaller models to think.
\citet{paliotta2025thinkingslowfastscaling} and \citet{liao-etal-2025-skintern} transfer reasoning ability from large models to small ones by distillation.
\citet{liu2025quantizationhurtsreasoningempirical} study the impact of quantization for reasoning LLMs.
Our approach leverages collaborative inference between large and small models, combining the small model's efficiency with the large model's accuracy to achieve an improved efficiency-accuracy trade-off in reasoning LLMs.

\textbf{Speculative Decoding}\quad
Autoregressive language models generate one token in one decoding step, which greatly limits its decoding efficiency.
Speculative decoding \citep{stern2018blockwise,xia2023speculative,leviathan2023fast,chen2023accelerating} adopts lightweight models for drafting and target models for parallel verification to generate multiple tokens in one decoding step, thus achieving lossless acceleration.
The draft model can be an independent small model \citep{leviathan2023fast,chen2023accelerating,spector2023accelerating,chen2023cascade}, a submodule of the target model \citep{cai2024medusa,zhang2023draft,li2024eagle}, or an external corpus \citep{he2023rest}.
Draft tokens that match the original output of the target model will be accepted under a strict verification strategy.
Most speculative decoding methods perform token-level drafting, while we borrow the idea of speculation to perform thought-level drafting to improve the reasoning efficiency of LLMs.

\section{Speculative CoT}
\label{sec:scot}
\begin{figure}[t]
    \centering
    \includegraphics[width=\textwidth]{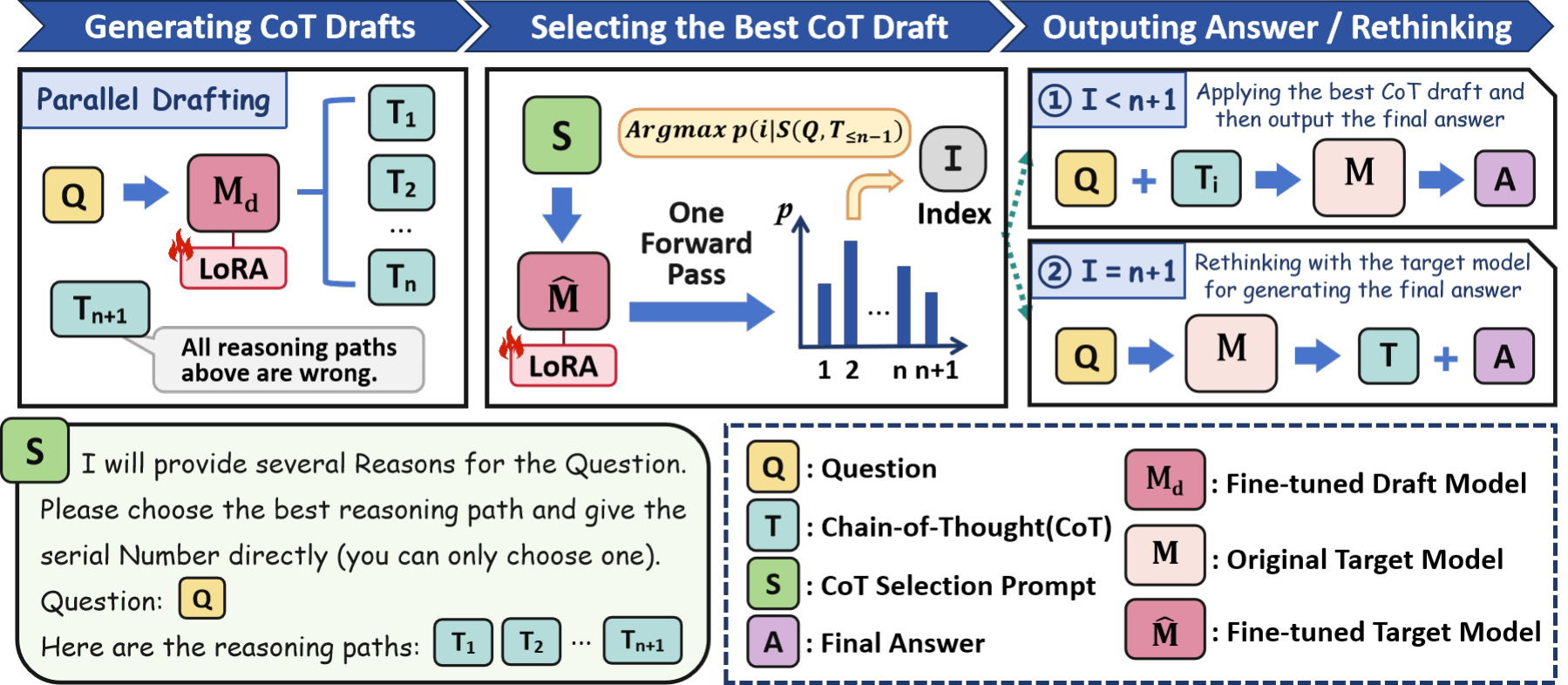}
    \caption{\label{fig:scot} Overview of Speculative Chain-of-Thought. Given a question $Q$, SCoT first applies a lightweight draft model to generate multiple CoT drafts in parallel. The draft model is fine-tuned with LoRA modules to align the thinking behavior of the target model. It appends a special CoT option for the case where all drafts are wrong. The target model for selecting the best CoT draft is also fine-tuned with LoRA modules for improved accuracy. With the designed prompt template $S$, only one forward propagation is needed to get the index of the best CoT. Once the best CoT is picked, SCoT directly adopts it for the original target model to generate the final answer. Only if no draft is selected, SCoT will rethink with the target model to ensure the quality of the generated answer.}
\end{figure}

This section introduces Speculative Chain-of-Thought (SCoT), an algorithm for accelerating LLM reasoning.
SCoT employs a small model to draft reasoning chains while utilizing the target model to select the best draft for generating the final answer.
The overview of SCoT is shown in Figure \ref{fig:scot}.

\subsection{Formulation}
For question $q$ in the given question set Q, the reasoning model $M$ first enters the thinking stage after inputting $q$. Specifically, M will output a chain-of-thought $T=(y_{1},y_{2},...,y_{l})$ starting and ending with pre-trained special tokens, where $y_{i}$ represents the $i$-th output token and $l$ is length of $T$. Take Deepseek-R1 \citep{deepseekai2025deepseekr1incentivizingreasoningcapability} as an example. The thinking process starts with ``\textit{<think>}'' and ends with ``\textit{</think>}''.
Next, the model generates the final answer $A$ based on $q$ and $T$.
Our goal is to reduce the inference overhead of $T$ while ensuring the quality of the final answer.

\subsection{Generating Drafts of Thoughts}

\begin{wrapfigure}{r}{0.36\textwidth}
    \vspace{-12pt}
    \centering
    \includegraphics[width=0.36\textwidth]{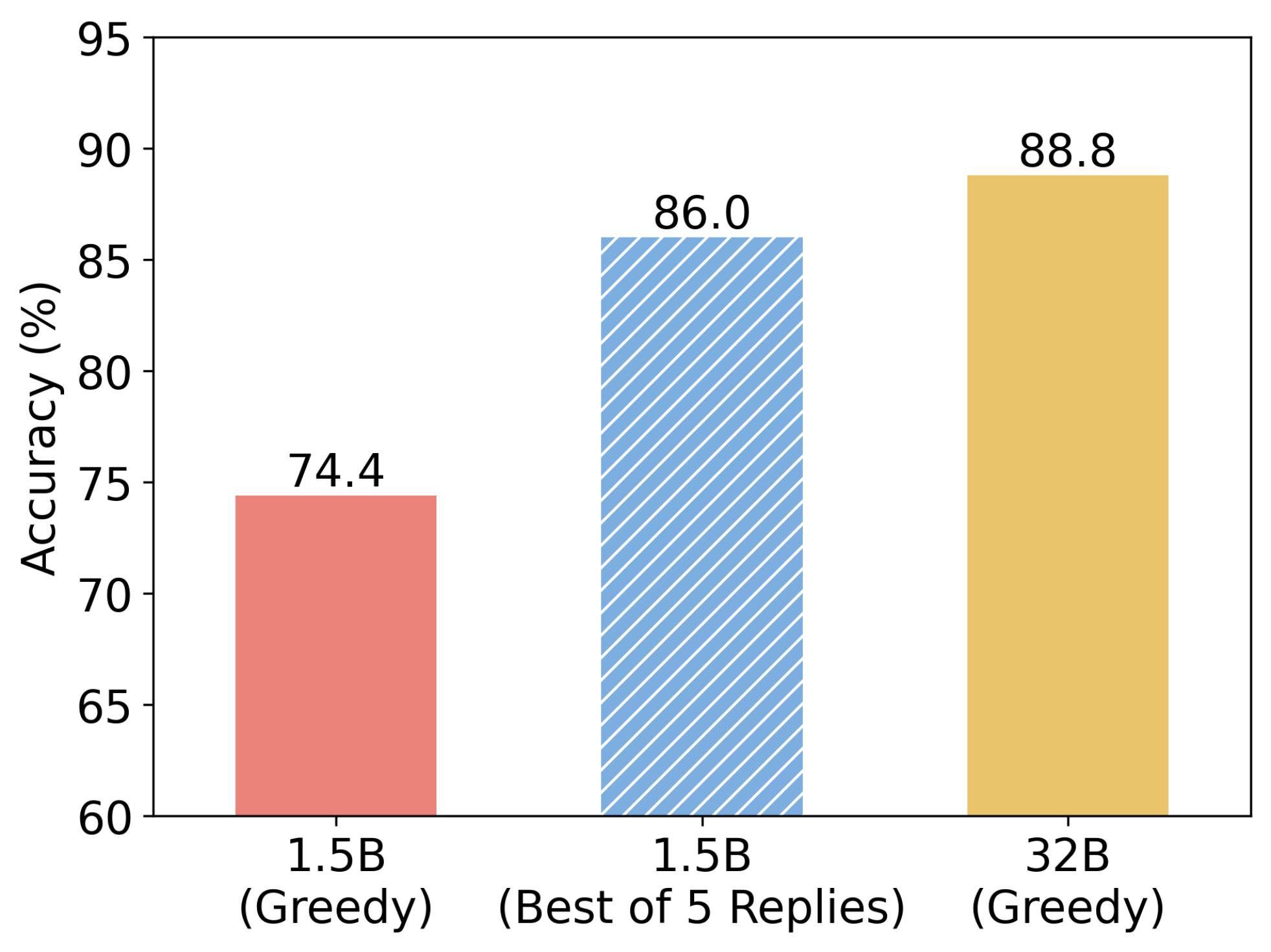}
    \caption{\label{fig:ac} Comparison of reasoning accuracy on GSM8K.}
    \vspace{-10pt}
\end{wrapfigure}

First, we start with a question: ``\textit{Can small models generate high-quality chain-of-thoughts?}''.
We sampled 500 samples in GSM8K \citep{cobbe2021gsm8k} to evaluate the reasoning accuracy of Deepseek-R1-Distill-Qwen-1.5B and Deepseek-R1-Distill-Qwen-32B \citep{deepseekai2025deepseekr1incentivizingreasoningcapability}.
Figure \ref{fig:ac} displays the results.
Under the greedy sampling method, the performance of the 1.5B model is significantly worse than that of the 32B model.
However, if we allow the 1.5B model to provide 5 answers under the nuclear sampling method and select the best one among the 5 replies for each sample, its overall accuracy on GSM8K is close to that of the 32B model.

From the above experiments, we can conclude that the small model has the potential to generate high-quality thought chains, but a stronger verifier is needed to control or select its generated results. Therefore, in order to accelerate the thinking process of reasoning LLMs, we first use a smaller model $M_{d}$ to draft $n$ chains of thoughts in parallel:
\begin{equation}
    [T_{d}^{1},T_{d}^{2},...,T_{d}^{n}]=M_{d}(\underset{n}{[\underbrace{q,q,...,q}]}).
\end{equation}
We use temperature sampling for the small model thinking to enhance the diversity of drafts of CoTs. Thanks to their smaller parameter size, smaller models can draft reasoning chains significantly faster than large models when the length of the CoT is the same, thus speeding up the thinking process.

\subsection{Thinking Behavior Alignment}
Although small models have higher throughput than large ones, significant differences in their reasoning behavior may still result in limited, or even reduced, reasoning speed.
In our experiments, we observed that even models of the same series can exhibit significant differences in thinking behavior.
The inference speed of Deepseek-R1-Distill-Qwen-1.5B is about 6 times that of Deepseek-R1-Distill-Qwen-32B. However, the average CoT length of the 1.5B model on the GSM8K dataset is more than 4 times that of the 32B model. The CoT of the 1.5B model contains more redundant tokens and repeated reflections.
Therefore, the draft model requires training in alignment of thinking behavior to ensure drafting efficiency.

We adopt 1500 samples from the GSM8K training set to make the target model generate the corresponding thought chains.
The training data $d=(x_{1},x_{2},...,x_{m},y_{1},y_{2},...,y_{l})$ is a combination of the question and generated CoT, where $m$ is the length of the question and $x_{ \leq m}$ are input tokens.
Then we use LoRA modules \citep{hu2022lora} to fine-tune the draft model with the cross-entropy loss:
\begin{equation}
    \mathcal{L}_{Draft} = -\frac{1}{l} \sum_{i=1}^{l}   \log p_{M_{d}}(y_{i}|x_{ \leq m},y_{<i} ),
\end{equation}
where $p_{M_{d}}(\cdot |\cdot)$ represents the output probability distribution of $M_{d}$.
The original parameters of the draft model are frozen during training.
The LoRA modules are adapted to the $\boldsymbol{Q}$ and $\boldsymbol{V}$ matrices in the attention block of each layer.
After training, we merge the LoRA module parameters into the weight matrices of the draft model. Therefore, there is no additional overhead during inference.

\subsection{Draft Selection and Error Correction}
After generating CoT drafts ($T_{1},T_{2},...,T_{n}$) by the fine-tuned draft model, we design a prompt template $S$ for selecting these thought chains.
Figure \ref{fig:scot} displays the specific content of this template.
For particularly challenging problems where all CoT drafts may be incorrect, we aim for the target model to detect such cases to ensure the accuracy of the final answer.
For this purpose, we introduce a special option $T_{n+1}$, which indicates that all CoT drafts are wrong.
Through a single forward propagation, we can efficiently obtain the CoT index with the highest probability from the output distribution of the next token by the target model:
\begin{equation}
    index = \underset{i \in \mathcal{V}}{\arg\max}\, P_{M}(i|S(q,T_{\leq n+1})),\,\mathcal{V}=\{1,2,...,n+1\}.
\end{equation}
To improve the accuracy of the target model in selecting the correct CoT drafts and detecting errors, we also use the LoRA modules to fine-tune the target model.
We use 500 samples from the GSM8K training set to construct the training data.
For each sample, we use the draft model to generate $n$ thought chains and deploy template $S$ as the input.
We judge the correctness of the CoT drafts based on the ground truth and construct the label set $\mathcal{Y}$, which is the set of indices for all correct drafts.
See Appendix \ref{sec:a1} for examples.
We design the following loss function to fine-tune the target model:
\begin{equation}
    \mathcal{L}_{Target} = \min \{-\log p_{M}(y|S(q,T_{\leq n+1}))|y\in \mathcal{Y}\}.
\end{equation}
Likewise, the original model parameters are frozen, and parameters in LoRA modules are merged into the weight matrices of $\boldsymbol{Q}$ and $\boldsymbol{V}$. The fine-tuned target model is only used for CoT selection.

There are two situations when selecting CoT. The first is that the output $index < n + 1$, that is, the target model selects the best one $T_{index}$ among the CoTs generated by the draft model.
In this case, we directly let the target model generate the final answer based on $q$ and $T_{index}$.
The second situation is the output $index = n + 1$, which indicates that the problem is too hard and the draft model failed to produce correct reasoning.
To maintain performance in the face of these complex problems, we instruct the target model to rethink the question and generate the final answer.

\section{Experiments}
\label{sec:ex}

\subsection{Settings}

We adopt Deepseek-R1-Distill-Qwen-32B \citep{deepseekai2025deepseekr1incentivizingreasoningcapability} and Deepseek-R1-Distill-Llama-70B (abbreviated as D-Qwen-32B and D-Llama-70B) as the target models.
We employ Deepseek-R1-Distill-Qwen-1.5B and Deepseek-R1-Distill-Llama-8B as the draft models for them, respectively.
We conducte the two groups of experiments under different hardware environments.
We use a single A100-PCIE-40GB GPU for the 1.5B model and 4 A100-PCIE-40GB GPUs for the 32B model.
While we deploy the 8B model on a single H800-80GB GPU and the 80B model on 4 H800-80GB GPUs.
We evaluate the efficiency and accuracy of Speculative CoT on reasoning datasets of various difficulty levels, including GSM8K \citep{cobbe2021gsm8k}, MATH \citep{hendrycks2021measuring}, GaoKao-En-2023 \citep{liao2024mariomathreasoningcode}, CollegeMath \citep{tang2024mathscalescalinginstructiontuning} and Olympiad \citep{he-etal-2024-olympiadbench}.
As stated in Section \ref{sec:scot}, we use data from GSM8K training set to fine-tune the models with LoRA modules.
Note that there is no overlap in the data used for training and evaluation. 

For SCoT, we set the rank of LoRA modules to 8 for D-Qwen-1.5B/32B and D-Llama-8B, and we set it to 16 for D-Llama-70B.
Learning rate is set to 5e-5 and 1e-4 for the draft models and target models, respectively.
The number of CoT drafts $n$ is set to 5. 
We set the temperature for sampling to 0.6 in CoT drafting process.
To avoid running out of memory when selecting the drafts, the maximum draft length is set to 5000.
We set the maximum generation length of CoT to 20480.
We compare the reasoning latency and accuracy of Speculative CoT with the vanilla reasoning by the target models.

We also compare the performance of SCoT with TokenSkip \citep{xia2025tokenskipcontrollablechainofthoughtcompression}, which trains the target model with compressed CoT data to shorten the mean CoT length. 
The original CoT data are generated by the original target model.
Then it compresses the thought chains through LLMLingua-2 \citep{pan2024llmlingua}.
We adopt the training set of GSM8K to construct the training data.
To compare the accuracy of the two methods at similar acceleration ratios, we set the compression ratio to 0.5 for TokenSkip, which means that about half of the tokens in thought chains are dropped.

\subsection{Main Results}
\label{sec:res}
\begin{table}[h!]
\captionsetup{skip=6pt}
\caption{Comparison of CoT length and reasoning efficiency. We report average CoT length of the target model $l_{M}$, average CoT length of the draft model $l_{M_{d}}$ and average CoT latency of each sample $t(\text{sec})$. Speed-up ratio $r$ is calculated by average thinking time compared with vanilla decoding.}
\label{tab:speed}
\renewcommand{\arraystretch}{1.2}
\resizebox{\textwidth}{!}{
\begin{tabular}{lcccccccccc}
\toprule
      \multirow{2}{*}{\textbf{Method}}             & \multicolumn{2}{c}{\textbf{GSM8K}}                                                                                                   & \multicolumn{2}{c}{\textbf{MATH}}      &    \multicolumn{2}{c}{\textbf{GaoKao}}      & \multicolumn{2}{c}{\textbf{CollegeMath}} &      \multicolumn{2}{c}{\textbf{Olympiad}}                                                                              \\ \cmidrule{2-11}
   & $l_{M}$ & $l_{M_{d}}$ & $l_{M}$ & $l_{M_{d}}$ & $l_{M}$ & $l_{M_{d}}$ & $l_{M}$ & $l_{M_{d}}$ & $l_{M}$ & $l_{M_{d}}$    \\
\midrule
D-Qwen-32B  &    302  &       -    &    2123   &    -  &  3230  & -   &  1051  & - & 8431  &  - \\
SCoT(32B+1.5B)    &   45    &   265   &   151  &   1855      &    630 &  2767   &    110   &  1136   & 3553 &  4194 \\ 
\hline
D-Llama-70B  &   326   &       -    &   1492  &    -  &  2521  & -   &  920 & - &  5503 &  - \\
SCoT(70B+8B)    &  47   &   410  &  70  &    2395  &    501  & 2986  &  120   & 1913   & 2579  & 4319 \\ 
\midrule
& $t$ & $r$ & $t$ & $r$  & $t$ & $r$ &$t$ & $r$ & $t$ & $r$  \\
\midrule
D-Qwen-32B  &    26.5  &       1.00    &    225.1   &   1.00  &  369.5  & 1.00  &  95.3  & 1.00  &  1092.2 &  1.00 \\
SCoT(32B+1.5B)    &   \textbf{11.7}    &   \textbf{2.26}   &   \textbf{77.0}  &   \textbf{2.92}      &    \textbf{160.1} &  \textbf{2.31}  &   \textbf{43.7}    &  \textbf{2.18}   & \textbf{575.7} &  \textbf{1.90} \\ 
\hline
D-Llama-70B  &   21.6   &       1.00    &   105.5  &    1.00  &  188.3  & 1.00   & 63.3   & 1.00 & 443.6  &  1.00 \\
SCoT(70B+8B)    &  \textbf{11.0}   &   \textbf{1.96}  &  \textbf{57.8}  &   \textbf{1.83}   &   \textbf{111.0}  &  \textbf{1.70}  &   \textbf{50.4}  &  \textbf{1.26}  &  \textbf{325.0} &  \textbf{1.36} \\ 

\bottomrule
\end{tabular}
}
\end{table}


\begin{figure}[h!]
    \centering
    \includegraphics[width=\textwidth]{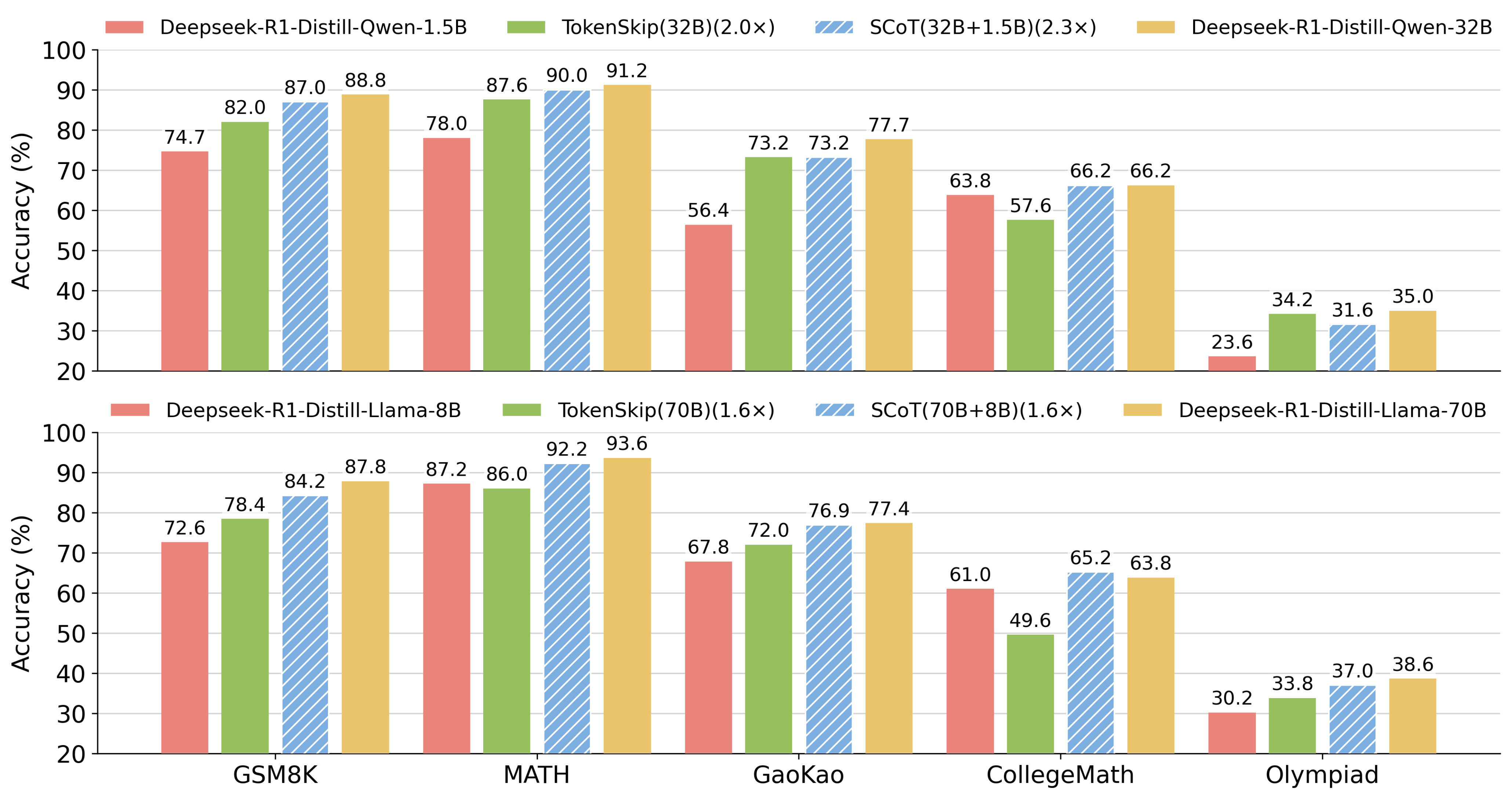}
    \caption{\label{fig:mb} Comparison of accuracy of the final answer on the 5 datasets with Deepseek-R1-Distill-Qwen (top) and Deepseek-R1-Distill-Llama (bottom). The compression ratio for TokenSkip is set to 0.5. The speed-up ratio in the legends represents the latency speed-up for TokenSkip and SCoT.}
    \vspace{-10pt}
\end{figure}

\textbf{Lower Reasoning Latency} \quad Table \ref{tab:speed} demonstrates the average CoT length and the average CoT latency of each sample on the 5 datasets.
$l_{M}$ and $l_{M_{d}}$ represent the numbers of tokens generated by the target model and the draft model during the thinking stage, respectively.
Since the draft model generates 5 draft chains in parallel, $l_{M_{d}}$ refers to the length of the longest draft CoT.
The average latency $t$ is measured in seconds.
For the 32B model, $l_{M}$ exhibits a wide range of 302 to 8431 across different datasets.
Except for MATH where the average CoT length of SCoT ($l_{M}+l_{M_{d}}$) is shorter than that of the 32B model, the overall CoT lengths of the two methods are close in the Qwen group.
However, SCoT generates more tokens during the reasoning phase in the Llama group, where the speed-up ratio is relatively lower.
For SCoT, most of the reasoning tokens are generated by the draft model, and the CoT tokens generated by the target model account for 2.8\%$\sim$45.9\% of the total CoT tokens.
As task difficulty increases, small models exhibit higher error rates, thus necessitating greater involvement from the target model in the reasoning process.
The draft selection mechanism dynamically determines whether the target model should reconsider its output, evaluating both problem complexity and draft generation quality. For simpler tasks, thought chains are predominantly produced by the draft model. Conversely, the proposed mechanism allocates more thought chain generation to the target model when handling complex problems.

The throughput of the draft model is about 5 times that of the corresponding target model for both groups in our experimental environment.
The CoT latency of SCoT is mainly composed of the target model reasoning time, draft model reasoning time, and draft selection time. 
For the GSM8K dataset in the Qwen group, the proportions are approximately 33.9\%, 60.2\%, and 5.9\% respectively.
With longer generated CoTs (e.g., on Olympiad), draft selection incurs minimal overhead.
Generally, SCoT achieves greater speed-up ratios on simpler tasks where the CoT drafts demonstrate higher quality.
SCoT achieves a speed-up ratio ranging from 1.26× to 2.92× compared to vanilla reasoning.

\textbf{Near-target-model Level Accuracy} \quad Figure \ref{fig:mb} displays the accuracy of different methods for Deepseek-R1-Distill-Qwen and Deepseek-R1-Distill-Llama on these datasets.
Comparative results indicate that SCoT maintains near-target-model performance levels on all datasets (even for the Olympiad-level task), with clearly superior accuracy compared to standalone draft model reasoning.
For CollegeMath, SCoT even achieves a better accuracy compared to the target model.
Moreover, SCoT outperforms TokenSkip in generation accuracy on most groups except Olympiad with the Qwen model, even with higher speed-up ratios, exhibiting better accuracy-efficiency trade-off.

\subsection{Ablation Studies}
\vspace{-10pt}
\begin{table}[h]
\centering
\captionsetup{skip=6pt}
\caption{Results for ablation study. Best results are in bold.}
\label{tab:ma}
\renewcommand{\arraystretch}{1.15}
\resizebox{\textwidth}{!}{
\begin{tabular}{lccccc}
\toprule
\textbf{Method}     & \textbf{GSM8K}                                     & \textbf{MATH}      &   \textbf{GaoKao}      & \textbf{CollegeMath} &      \textbf{Olympiad}     \\                        
\midrule
SCoT (32B+1.5B)  &  \textbf{87.0} & \textbf{90.0} & \textbf{73.2} & \textbf{66.2} & \textbf{31.6} \\
\quad w Single Draft   & 85.6 & 86.6 & 69.9 & 63.8 & 30.0 \\ 
\quad w/o Tuning $M$   & 85.6 & \textbf{90.0} & 69.9 & 64.8 & 30.0 \\ 
\quad w/o Error Correction   & 85.6 & \textbf{90.0} & 71.4 & 65.6 & 30.2 \\ 

\bottomrule
\end{tabular}
}
\end{table}
We conduct an ablation study in this section.
We carry out 3 sets of experiments to observe the changes in accuracy on the 5 datasets with Deepseek-R1-Distill-Qwen: 1) SCoT with Single Draft: Only one CoT draft is generated for each sample (except for the special draft $T_{n+1}$), 2) SCoT without Tuning $M$: We use the original target model instead of the fine-tuned version to select the drafts, 3) SCoT without Error Correction: We remove $T_{n+1}$ and the rethinking strategy.
Table \ref{tab:ma} shows the results.
SCoT exhibits the highest accuracy on all datasets, indicating that both of the three proposed techniques contribute to the reasoning performance.
Drafting multiple thought chains increases the upper limit of draft quality, thus enhancing the potential for higher-quality outputs.
In preliminary experiments, we found that without fine-tuning, $M$ never selects the ``\textit{All reasoning paths are wrong.}'' option for any CoT draft candidates.
After training, the target model can acquire a certain error correction capability.
We will discuss this further in Section \ref{sec:select}.

\begin{table}[h]
\centering
\captionsetup{skip=6pt}
\caption{Comparative analysis of average CoT draft length on DeepSeek-R1-Distill-Qwen-1.5B and Deepseek-R1-Distill-Llama-8B with versus without thinking behavior alignment training.}
\label{tab:ablation}
\renewcommand{\arraystretch}{1.2}
\resizebox{\textwidth}{!}{
\begin{tabular}{lccccc}
\toprule
\textbf{Method}     & \textbf{GSM8K}                                     & \textbf{MATH}      &   \textbf{GaoKao}      & \textbf{CollegeMath} &      \textbf{Olympiad}     \\                        
\midrule
Deepseek-R1-Distill-Qwen-1.5B  &  606 & 2092 & 2589 & 1355 & 4032 \\
\textbf{+ Thinking Behavior Alignment}   & \textbf{193}  & \textbf{1183} & \textbf{1487} & \textbf{709} & \textbf{3554} \\ 
\hline
Deepseek-R1-Distill-Llama-8B  &  585 & 1979 & 3223 & 1285 & 6474 \\
\textbf{+ Thinking Behavior Alignment}   & \textbf{258} & \textbf{1536} & \textbf{2253} & \textbf{1254} & \textbf{3783} \\

\bottomrule
\end{tabular}
}
\end{table}
In order to verify the effectiveness of thinking behavior alignment, we compare the average CoT draft length generated by the original draft models and the fine-tuned versions.
Table \ref{tab:ablation} shows the mean CoT length on the 5 datasets.
The CoT length generated by the fine-tuned model is significantly shorter than that of the original model on all datasets for both the two draft models, especially for GSM8K.
Therefore, thinking behavior alignment improves the efficiency of the drafting process.

\subsection{Comparison with Speculative Decoding}
\vspace{-10pt}
\begin{table}[ht]
\captionsetup{skip=6pt}
\caption{Comparison of reasoning efficiency with auto-regressive decoding (ARD) and speculative decoding (SD) on the five datasets with the two models. $s$ represents the average throughput during the reasoning phase (including the prefilling phase). For SCoT, only valid tokens (generated by either the draft model or the target model) are included in the calculation of $s$. $r^{\prime}$ represents the speed-up ratio of throughput. $r$ stands for the speed-up ratio of average reasoning latency of each sample.}
\label{tab:sd}
\renewcommand{\arraystretch}{1.25}
\resizebox{\textwidth}{!}{
\begin{tabular}{l|ccc|ccc|ccc|ccc|ccc}
\toprule
 \multicolumn{1}{l}{\multirow{2}{*}{\textbf{Method}}}           & \multicolumn{3}{c}{\textbf{GSM8K}}                                                                                                   & \multicolumn{3}{c}{\textbf{MATH}}      &    \multicolumn{3}{c}{\textbf{GaoKao}}      & \multicolumn{3}{c}{\textbf{CollegeMath}} &      \multicolumn{3}{c}{\textbf{Olympiad}}                                                                              \\ \cmidrule{2-16}
& $s$ & $r^{\prime}$ & $r$ & $s$ & $r^{\prime}$ & $r$ & $s$ & $r^{\prime}$ & $r$ & $s$ & $r^{\prime}$ & $r$ & $s$ & $r^{\prime}$ & $r$  \\
\midrule
\multicolumn{16}{c}{\textit{Deepseek-R1-Distill-
Qwen-32B (A100-PCIE-40GB)}} \\
\midrule
ARD  & 11.4 & 1.00  &  1.00  & 9.4 & 1.00  &  1.00  & 8.7 & 1.00  &  1.00 & 11.0 & 1.00  &  1.00 & 7.9 & 1.00  &  1.00 \\
SD & 18.5 & 1.63  &  1.87  & 13.4 & 1.42  &  1.41  &  12.1 & 1.38  &  1.32 & 15.6 & 1.42  &  1.34 & 9.1 & 1.16  &  0.59 \\
SCoT  & \textbf{23.9} & \textbf{2.10}  &  \textbf{2.26}  & \textbf{25.4} & \textbf{2.69}  &  \textbf{2.92}  & \textbf{19.1} & \textbf{2.18}  &  \textbf{2.31} & \textbf{27.9} & \textbf{2.53}  &  \textbf{2.18} & \textbf{11.2} & \textbf{1.42}  &  \textbf{1.90} \\
\midrule
\multicolumn{16}{c}{\textit{Deepseek-R1-Distill-
Llama-70B (H800-80GB)}} \\
\midrule
ARD  & 14.5 & 1.00  &  1.00  & 14.1 & 1.00  &  1.00  & 13.4 & 1.00  &  1.00 & 14.5 & 1.00  &  1.00 & 12.7 & 1.00  &  1.00 \\
SD & 25.1 & 1.64  &  1.63  & 21.2 & 1.50  &  1.49  & 17.6 & 1.31  &  1.35 & 23.2 & 1.59  &  \textbf{1.61} & 12.6 & 0.99  &  0.58 \\
SCoT   & \textbf{37.3} & \textbf{2.45}  &  \textbf{1.96}  & \textbf{41.8} & \textbf{2.96}  &  \textbf{1.83}  & \textbf{28.7} & \textbf{2.14}  &  \textbf{1.70} & \textbf{37.5} & \textbf{2.58}  &  1.26 & \textbf{17.4} & \textbf{1.37}  &  \textbf{1.36} \\

\bottomrule
\end{tabular}
}
\end{table}

We compare the reasoning efficiency of SCoT with speculative decoding with large and small models collaboration \citep{leviathan2023fast,kim2023speculative} in this section.
Speculative decoding generates several tokens with the draft model and then verifies the candidates with the target model in each decoding step.
Tokens that match the original output will be accepted during verification.
We evaluate the throughput and latency with Deepseek-R1-Distill- Qwen-32B and Deepseek-R1-Distill-Llama-70B.
We use the original Deepseek-R1-Distill-Qwen-1.5B and Deepseek-R1-Distill-Llama-8B as the draft models for speculative decoding.
The maximum draft length is set to 5.

Table \ref{tab:sd} displays the results.
The throughput of auto-regressive decoding is affected by the length of the sequence.
Speculative decoding performs well when the sequence length is short. 
It achieves a reasoning latency speed-up ratio of up to 1.87 and 1.63 for the Qwen model and the Llama model, respectively.
However, for the Olympiad dataset, where the average generation length exceeds 5000 tokens, the drafting overhead grows with sequence length, ultimately causing deceleration for both the target models.
Note that speculative decoding may produce outputs that deviate from the original model when using half-precision. In certain cases (e.g., GSM8K dataset with the Qwen model), the shorter generation lengths can lead to $r^{\prime}<r$.
Employing the same draft models, SCoT achieves a higher throughput on all datasets.
It shows a speed-up ratio of up to 2.96 for throughput.
It also exhibits a better acceleration for reasoning latency (except for CollegeMath in the Llama group).
While speculative decoding requires multiple verifications, SCoT's single draft-verify cycle significantly reduces the frequent model reloads between GPU high-bandwidth memory (HBM) and on-chip cache.
Therefore, generation length exhibits a relatively minor impact on SCoT's efficiency (as stated in Section \ref{sec:res} the speed-up ratio of SCoT is mainly affected by the difficulty of the task), reflecting its good adaptability to long sequence generation scenarios.

\subsection{Draft Selection Accuracy Analysis}
\label{sec:select}
\begin{wraptable}[9]{r}{4.85cm}
\vspace{-12pt}
\captionsetup{skip=6pt}
\caption{The number of two types of CoT drafts and the accuracy of draft selection on GSM8K.}
\label{tab:selection}
\renewcommand{\arraystretch}{1.1}
\begin{tabular}{lcc}
\toprule
\textbf{Type} & \textbf{Class 1} & \textbf{Class 2} \\
\midrule
\textbf{Amount} & 450 & 50 \\
\textbf{Accuracy} & 85.1 & 52.0 \\
\bottomrule
\end{tabular}
\end{wraptable}
In this section, we study the draft selection and error detection accuracy of the fine-tuned  Deepseek-R1-Distill-Qwen-32B on the 500 samples in the test set of GSM8K.
We divide these samples into two categories according to their labels. Samples for which the draft model can generate the correct reasoning chain form Class 1 (the label $y \neq n+1$), while others constitute Class 2 (the label $y = n+1$).
Table \ref{tab:selection} shows the statistical results.
There are 450 first-class samples out of the 500 samples.
Since the second type of sample is more difficult to predict and the training data is limited, the target model demonstrates more accurate prediction for Class 1 samples than Class 2 samples.
Our preliminary tests show that direct prediction by the original target model on the second-class samples yields zero accuracy.
After fine-tuning, the model can correctly predict the labels for about half of the Class 2 samples.

\subsection{Impact on Length of the Final Answer}

We study the impact of SCoT on the length of the final answer in this section.
Table \ref{tab:answer} displays the answer length of SCoT on the 5 datasets.
In addition to reducing the latency of the reasoning phase, SCoT also reduces the length of the generated final answer compared to the target model and even brings additional inference efficiency improvements in the Qwen group.
For the Llama group, SCoT generates final answers with lengths comparable to those produced by the original target model.
\begin{table}[h]
\centering
\captionsetup{skip=6pt}
\caption{Average length of the final answer generated by Deepseek-R1-Distill-Qwen-32B and SCoT.}
\label{tab:answer}
\renewcommand{\arraystretch}{1.2}
\resizebox{\textwidth}{!}{
\begin{tabular}{lccccc}
\toprule
\textbf{Method}     & \textbf{GSM8K}                                     & \textbf{MATH}      &   \textbf{GaoKao}      & \textbf{CollegeMath} &      \textbf{Olympiad}     \\                        
\midrule
Deepseek-R1-Distill-Qwen-32B  &  314 & 632 & 948 & 416 & 2952 \\
SCoT (32B+1.5B)   & 270  & 358 & 610 & 357 & 2179 \\ 
\hline
Deepseek-R1-Distill-Llama-70B  &  238 & 256 & 453 & 312 & 840 \\
SCoT (70B+8B)   & 212  & 310 & 568 & 314 & 1068 \\ 
\bottomrule
\end{tabular}
}
\end{table}

\subsection{Performance under Different Number of Chains}
\label{sec:chain}
\begin{wrapfigure}{r}{0.46\textwidth}
    \vspace{-12pt}
    \centering
    \includegraphics[width=0.46\textwidth]{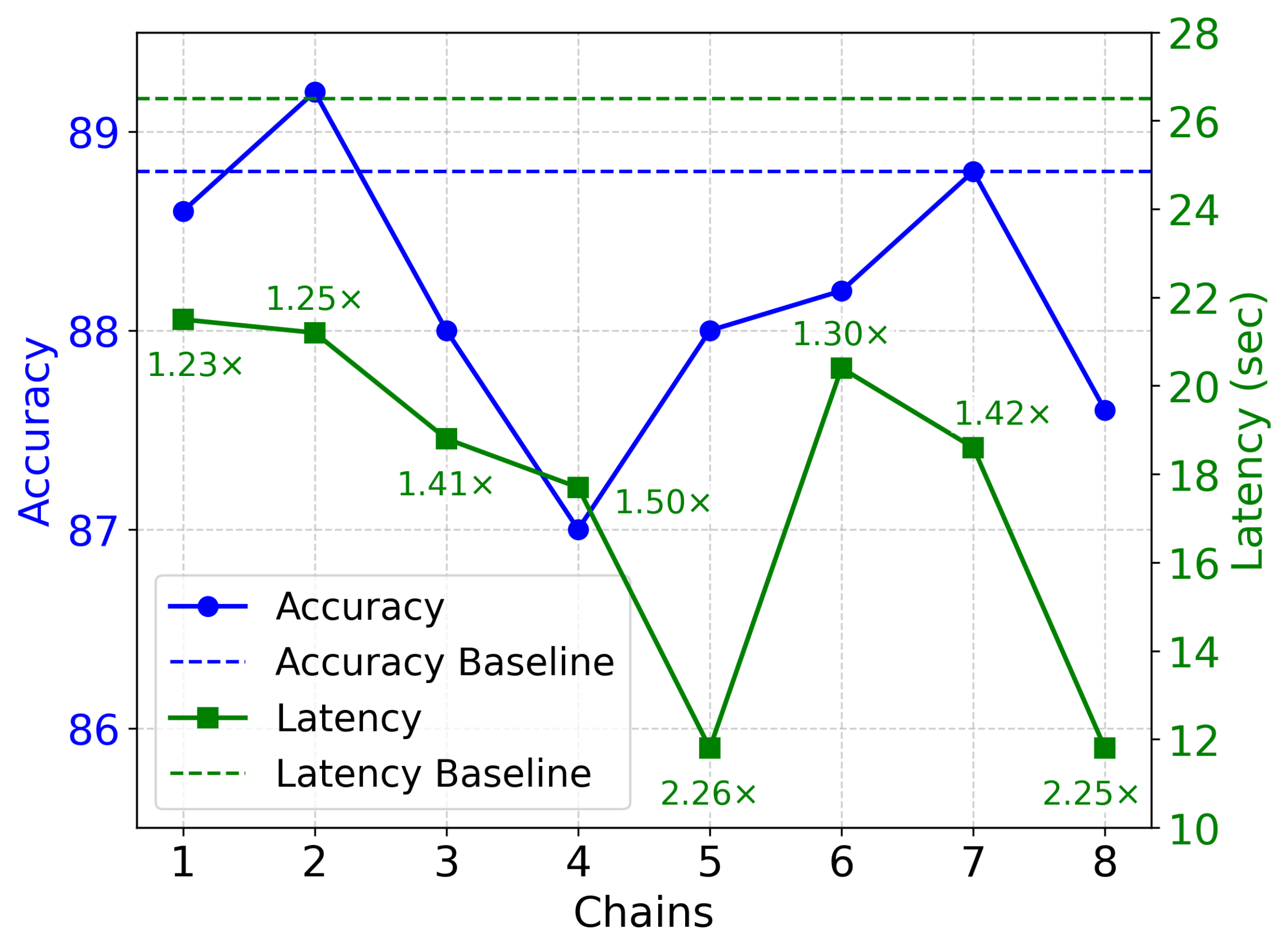}
    \caption{\label{fig:chain} Impact of the chain number on reasoning accuracy and latency on GSM8K.}
    \vspace{-10pt}
\end{wrapfigure}
To explore the best number of CoT draft candidates, we evaluate the reasoning accuracy and latency under different numbers of draft thought chains in this section.
We conduct an experiment with the Deepseek-R1-Distill-Qwen models on the GSM8K dataset.
We tested the performance when drafting 1 to 8 thought chains.
Figure \ref{fig:chain} displays the results.
The blue and green lines represent the accuracy and reasoning latency, respectively.
The two dotted lines show the baseline accuracy and latency of Deepseek-R1-Distill-Qwen-32B.
Increasing the number of chains increases the probability of including high-quality reasoning paths in the set of drafts.
When using 2 draft chains, while the latency reduction remains modest, SCoT surprisingly outperforms the target model in accuracy, demonstrating the remarkable potential of collaborative model generation.
A higher speed-up ratio indicates that SCoT accepts more reasoning chains generated by the draft model.
SCoT can reduce its reliance on the target model due to the increase in the number of candidates, thus achieving a higher speed-up ratio.
However, as the number of chains increases, the difficulty of training the target model to select the correct draft also increases.
Meanwhile, due to the inherent randomness in training procedures and the limited scale of training data, it exhibits fluctuations in reasoning accuracy.
Moreover, an oversupply of chains complicates error detection for the target model, which may lead to accuracy degradation despite the improved speed.
Overall, SCoT demonstrates optimal acceleration performance with 5 thought chains while maintaining acceptable accuracy degradation.

\section{Conclusion}
In this paper, we proposed a novel reasoning framework called Speculative Chain-of-Thought to alleviate the latency bottleneck of reasoning LLMs.
The proposed method leverages the efficiency of small models for parallel CoT drafting and the higher accuracy of the large model for drafting selection and error correction.
The draft selection mechanism adaptively decides whether the target model needs to rethink based on the difficulty of the problem and the quality of the draft generation.
Thought chains are mostly generated by the draft model for simple tasks.
While more thought chains will be allocated to the target model to generate for complex problems.
Speculative Chain-of-Thought demonstrates a speed-up ratio of up to 2.92× for the reasoning process while achieving near-target-model-level accuracy on 5 datasets of different difficulties.
It exhibits the potential for future work to realize a better accuracy-efficiency trade-off for reasoning LLMs through model collaboration.

\clearpage
\medskip
\bibliographystyle{apalike}
\bibliography{reference}

\begin{thebibliography}{}

\bibitem[Aytes et~al., 2025]{aytes2025sketch}
Aytes, S.~A., Baek, J., and Hwang, S.~J. (2025).
\newblock Sketch-of-thought: Efficient llm reasoning with adaptive cognitive-inspired sketching.
\newblock {\em arXiv preprint arXiv:2503.05179}.

\bibitem[Cai et~al., 2024]{cai2024medusa}
Cai, T., Li, Y., Geng, Z., Peng, H., Lee, J.~D., Chen, D., and Dao, T. (2024).
\newblock Medusa: Simple llm inference acceleration framework with multiple decoding heads.
\newblock {\em arXiv preprint arXiv:2401.10774}.

\bibitem[Chen et~al., 2023a]{chen2023accelerating}
Chen, C., Borgeaud, S., Irving, G., Lespiau, J.-B., Sifre, L., and Jumper, J. (2023a).
\newblock Accelerating large language model decoding with speculative sampling.
\newblock {\em arXiv preprint arXiv:2302.01318}.

\bibitem[Chen et~al., 2023b]{chen2023cascade}
Chen, Z., Yang, X., Lin, J., Sun, C., Huang, J., and Chang, K. C.-C. (2023b).
\newblock Cascade speculative drafting for even faster llm inference.
\newblock {\em arXiv preprint arXiv:2312.11462}.

\bibitem[Cobbe et~al., 2021]{cobbe2021gsm8k}
Cobbe, K., Kosaraju, V., Bavarian, M., Chen, M., Jun, H., Kaiser, L., Plappert, M., Tworek, J., Hilton, J., Nakano, R., Hesse, C., and Schulman, J. (2021).
\newblock Training verifiers to solve math word problems.
\newblock {\em arXiv preprint arXiv:2110.14168}.

\bibitem[DeepSeek-AI et~al., 2025]{deepseekai2025deepseekr1incentivizingreasoningcapability}
DeepSeek-AI, Guo, D., Yang, D., Zhang, H., Song, J., Zhang, R., Xu, R., Zhu, Q., Ma, S., Wang, P., Bi, X., Zhang, X., Yu, X., Wu, Y., Wu, Z.~F., Gou, Z., Shao, Z., Li, Z., Gao, Z., Liu, A., Xue, B., Wang, B., Wu, B., Feng, B., Lu, C., Zhao, C., Deng, C., Zhang, C., Ruan, C., Dai, D., Chen, D., Ji, D., Li, E., Lin, F., Dai, F., Luo, F., Hao, G., Chen, G., Li, G., Zhang, H., Bao, H., Xu, H., Wang, H., Ding, H., Xin, H., Gao, H., Qu, H., Li, H., Guo, J., Li, J., Wang, J., Chen, J., Yuan, J., Qiu, J., Li, J., Cai, J.~L., Ni, J., Liang, J., Chen, J., Dong, K., Hu, K., Gao, K., Guan, K., Huang, K., Yu, K., Wang, L., Zhang, L., Zhao, L., Wang, L., Zhang, L., Xu, L., Xia, L., Zhang, M., Zhang, M., Tang, M., Li, M., Wang, M., Li, M., Tian, N., Huang, P., Zhang, P., Wang, Q., Chen, Q., Du, Q., Ge, R., Zhang, R., Pan, R., Wang, R., Chen, R.~J., Jin, R.~L., Chen, R., Lu, S., Zhou, S., Chen, S., Ye, S., Wang, S., Yu, S., Zhou, S., Pan, S., Li, S.~S., Zhou, S., Wu, S., Ye, S., Yun, T., Pei, T., Sun, T., Wang, T., Zeng, W.,
  Zhao, W., Liu, W., Liang, W., Gao, W., Yu, W., Zhang, W., Xiao, W.~L., An, W., Liu, X., Wang, X., Chen, X., Nie, X., Cheng, X., Liu, X., Xie, X., Liu, X., Yang, X., Li, X., Su, X., Lin, X., Li, X.~Q., Jin, X., Shen, X., Chen, X., Sun, X., Wang, X., Song, X., Zhou, X., Wang, X., Shan, X., Li, Y.~K., Wang, Y.~Q., Wei, Y.~X., Zhang, Y., Xu, Y., Li, Y., Zhao, Y., Sun, Y., Wang, Y., Yu, Y., Zhang, Y., Shi, Y., Xiong, Y., He, Y., Piao, Y., Wang, Y., Tan, Y., Ma, Y., Liu, Y., Guo, Y., Ou, Y., Wang, Y., Gong, Y., Zou, Y., He, Y., Xiong, Y., Luo, Y., You, Y., Liu, Y., Zhou, Y., Zhu, Y.~X., Xu, Y., Huang, Y., Li, Y., Zheng, Y., Zhu, Y., Ma, Y., Tang, Y., Zha, Y., Yan, Y., Ren, Z.~Z., Ren, Z., Sha, Z., Fu, Z., Xu, Z., Xie, Z., Zhang, Z., Hao, Z., Ma, Z., Yan, Z., Wu, Z., Gu, Z., Zhu, Z., Liu, Z., Li, Z., Xie, Z., Song, Z., Pan, Z., Huang, Z., Xu, Z., Zhang, Z., and Zhang, Z. (2025).
\newblock Deepseek-r1: Incentivizing reasoning capability in llms via reinforcement learning.

\bibitem[Ding et~al., 2024]{ding2024breakchainlargelanguage}
Ding, M., Liu, H., Fu, Z., Song, J., Xie, W., and Zhang, Y. (2024).
\newblock Break the chain: Large language models can be shortcut reasoners.

\bibitem[Hao et~al., 2024]{hao2024traininglargelanguagemodels}
Hao, S., Sukhbaatar, S., Su, D., Li, X., Hu, Z., Weston, J., and Tian, Y. (2024).
\newblock Training large language models to reason in a continuous latent space.

\bibitem[He et~al., 2024]{he-etal-2024-olympiadbench}
He, C., Luo, R., Bai, Y., Hu, S., Thai, Z., Shen, J., Hu, J., Han, X., Huang, Y., Zhang, Y., Liu, J., Qi, L., Liu, Z., and Sun, M. (2024).
\newblock {O}lympiad{B}ench: A challenging benchmark for promoting {AGI} with olympiad-level bilingual multimodal scientific problems.
\newblock In Ku, L.-W., Martins, A., and Srikumar, V., editors, {\em Proceedings of the 62nd Annual Meeting of the Association for Computational Linguistics (Volume 1: Long Papers)}, pages 3828--3850, Bangkok, Thailand. Association for Computational Linguistics.

\bibitem[He et~al., 2023]{he2023rest}
He, Z., Zhong, Z., Cai, T., Lee, J.~D., and He, D. (2023).
\newblock Rest: Retrieval-based speculative decoding.
\newblock {\em arXiv preprint arXiv:2311.08252}.

\bibitem[Hendrycks et~al., 2021]{hendrycks2021measuring}
Hendrycks, D., Burns, C., Kadavath, S., Arora, A., Basart, S., Tang, E., Song, D., and Steinhardt, J. (2021).
\newblock Measuring mathematical problem solving with the {MATH} dataset.
\newblock In {\em Thirty-fifth Conference on Neural Information Processing Systems Datasets and Benchmarks Track (Round 2)}.

\bibitem[Hou et~al., 2025]{hou2025thinkprunepruninglongchainofthought}
Hou, B., Zhang, Y., Ji, J., Liu, Y., Qian, K., Andreas, J., and Chang, S. (2025).
\newblock Thinkprune: Pruning long chain-of-thought of llms via reinforcement learning.

\bibitem[Hu et~al., 2022]{hu2022lora}
Hu, E.~J., yelong shen, Wallis, P., Allen-Zhu, Z., Li, Y., Wang, S., Wang, L., and Chen, W. (2022).
\newblock Lo{RA}: Low-rank adaptation of large language models.
\newblock In {\em International Conference on Learning Representations}.

\bibitem[Jaech et~al., 2024]{jaech2024openai}
Jaech, A., Kalai, A., Lerer, A., Richardson, A., El-Kishky, A., Low, A., Helyar, A., Madry, A., Beutel, A., Carney, A., et~al. (2024).
\newblock Openai o1 system card.
\newblock {\em arXiv preprint arXiv:2412.16720}.

\bibitem[Kim et~al., 2023]{kim2023speculative}
Kim, S., Mangalam, K., Moon, S., Malik, J., Mahoney, M.~W., Gholami, A., and Keutzer, K. (2023).
\newblock Speculative decoding with big little decoder.
\newblock In {\em Thirty-seventh Conference on Neural Information Processing Systems}.

\bibitem[Leviathan et~al., 2023]{leviathan2023fast}
Leviathan, Y., Kalman, M., and Matias, Y. (2023).
\newblock Fast inference from transformers via speculative decoding.
\newblock In {\em International Conference on Machine Learning}, pages 19274--19286. PMLR.

\bibitem[Li et~al., 2024]{li2024eagle}
Li, Y., Wei, F., Zhang, C., and Zhang, H. (2024).
\newblock Eagle: Speculative sampling requires rethinking feature uncertainty.
\newblock {\em arXiv preprint arXiv:2401.15077}.

\bibitem[Liao et~al., 2025]{liao-etal-2025-skintern}
Liao, H., He, S., Hao, Y., Li, X., Zhang, Y., Zhao, J., and Liu, K. (2025).
\newblock {SKI}ntern: Internalizing symbolic knowledge for distilling better {C}o{T} capabilities into small language models.
\newblock In Rambow, O., Wanner, L., Apidianaki, M., Al-Khalifa, H., Eugenio, B.~D., and Schockaert, S., editors, {\em Proceedings of the 31st International Conference on Computational Linguistics}, pages 3203--3221, Abu Dhabi, UAE. Association for Computational Linguistics.

\bibitem[Liao et~al., 2024]{liao2024mariomathreasoningcode}
Liao, M., Luo, W., Li, C., Wu, J., and Fan, K. (2024).
\newblock Mario: Math reasoning with code interpreter output -- a reproducible pipeline.

\bibitem[Liu et~al., 2025]{liu2025quantizationhurtsreasoningempirical}
Liu, R., Sun, Y., Zhang, M., Bai, H., Yu, X., Yu, T., Yuan, C., and Hou, L. (2025).
\newblock Quantization hurts reasoning? an empirical study on quantized reasoning models.

\bibitem[Liu et~al., 2024]{liucan}
Liu, T., Guo, Q., Hu, X., Jiayang, C., Zhang, Y., Qiu, X., and Zhang, Z. (2024).
\newblock Can language models learn to skip steps?
\newblock In {\em The Thirty-eighth Annual Conference on Neural Information Processing Systems}.

\bibitem[Luo et~al., 2025]{luo2025o1prunerlengthharmonizingfinetuningo1like}
Luo, H., Shen, L., He, H., Wang, Y., Liu, S., Li, W., Tan, N., Cao, X., and Tao, D. (2025).
\newblock O1-pruner: Length-harmonizing fine-tuning for o1-like reasoning pruning.

\bibitem[Paliotta et~al., 2025]{paliotta2025thinkingslowfastscaling}
Paliotta, D., Wang, J., Pagliardini, M., Li, K.~Y., Bick, A., Kolter, J.~Z., Gu, A., Fleuret, F., and Dao, T. (2025).
\newblock Thinking slow, fast: Scaling inference compute with distilled reasoners.

\bibitem[Pan et~al., 2024]{pan2024llmlingua}
Pan, Z., Wu, Q., Jiang, H., Xia, M., Luo, X., Zhang, J., Lin, Q., R{\"u}hle, V., Yang, Y., Lin, C.-Y., et~al. (2024).
\newblock Llmlingua-2: Data distillation for efficient and faithful task-agnostic prompt compression.
\newblock In {\em Findings of the Association for Computational Linguistics ACL 2024}, pages 963--981.

\bibitem[Saunshi et~al., 2025]{saunshi2025reasoning}
Saunshi, N., Dikkala, N., Li, Z., Kumar, S., and Reddi, S.~J. (2025).
\newblock Reasoning with latent thoughts: On the power of looped transformers.
\newblock In {\em The Thirteenth International Conference on Learning Representations}.

\bibitem[Spector and Re, 2023]{spector2023accelerating}
Spector, B. and Re, C. (2023).
\newblock Accelerating llm inference with staged speculative decoding.
\newblock {\em arXiv preprint arXiv:2308.04623}.

\bibitem[Stern et~al., 2018]{stern2018blockwise}
Stern, M., Shazeer, N., and Uszkoreit, J. (2018).
\newblock Blockwise parallel decoding for deep autoregressive models.
\newblock {\em Advances in Neural Information Processing Systems}, 31.

\bibitem[Tang et~al., 2024]{tang2024mathscalescalinginstructiontuning}
Tang, Z., Zhang, X., Wang, B., and Wei, F. (2024).
\newblock Mathscale: Scaling instruction tuning for mathematical reasoning.

\bibitem[Wei et~al., 2023]{wei2023chainofthoughtpromptingelicitsreasoning}
Wei, J., Wang, X., Schuurmans, D., Bosma, M., Ichter, B., Xia, F., Chi, E., Le, Q., and Zhou, D. (2023).
\newblock Chain-of-thought prompting elicits reasoning in large language models.

\bibitem[Xia et~al., 2023]{xia2023speculative}
Xia, H., Ge, T., Wang, P., Chen, S.-Q., Wei, F., and Sui, Z. (2023).
\newblock Speculative decoding: Exploiting speculative execution for accelerating seq2seq generation.
\newblock In {\em Findings of the Association for Computational Linguistics: EMNLP 2023}, pages 3909--3925.

\bibitem[Xia et~al., 2025]{xia2025tokenskipcontrollablechainofthoughtcompression}
Xia, H., Li, Y., Leong, C.~T., Wang, W., and Li, W. (2025).
\newblock Tokenskip: Controllable chain-of-thought compression in llms.

\bibitem[Xu et~al., 2025a]{xu2025chaindraftthinkingfaster}
Xu, S., Xie, W., Zhao, L., and He, P. (2025a).
\newblock Chain of draft: Thinking faster by writing less.

\bibitem[Xu et~al., 2025b]{xu2025softcotsoftchainofthoughtefficient}
Xu, Y., Guo, X., Zeng, Z., and Miao, C. (2025b).
\newblock Softcot: Soft chain-of-thought for efficient reasoning with llms.

\bibitem[Yu et~al., 2024]{yu2024distilling21}
Yu, P., Xu, J., Weston, J., and Kulikov, I. (2024).
\newblock Distilling system 2 into system 1.

\bibitem[Zhang et~al., 2023]{zhang2023draft}
Zhang, J., Wang, J., Li, H., Shou, L., Chen, K., Chen, G., and Mehrotra, S. (2023).
\newblock Draft \& verify: Lossless large language model acceleration via self-speculative decoding.
\newblock {\em arXiv preprint arXiv:2309.08168}.

\bibitem[Zhang et~al., 2025]{zhang2025lightthinkerthinkingstepbystepcompression}
Zhang, J., Zhu, Y., Sun, M., Luo, Y., Qiao, S., Du, L., Zheng, D., Chen, H., and Zhang, N. (2025).
\newblock Lightthinker: Thinking step-by-step compression.

\end{thebibliography}

\clearpage
\appendix

\section{Sample Presentation}
\label{sec:a1}
\begin{figure}[h]
    \centering
    \includegraphics[width=\textwidth]{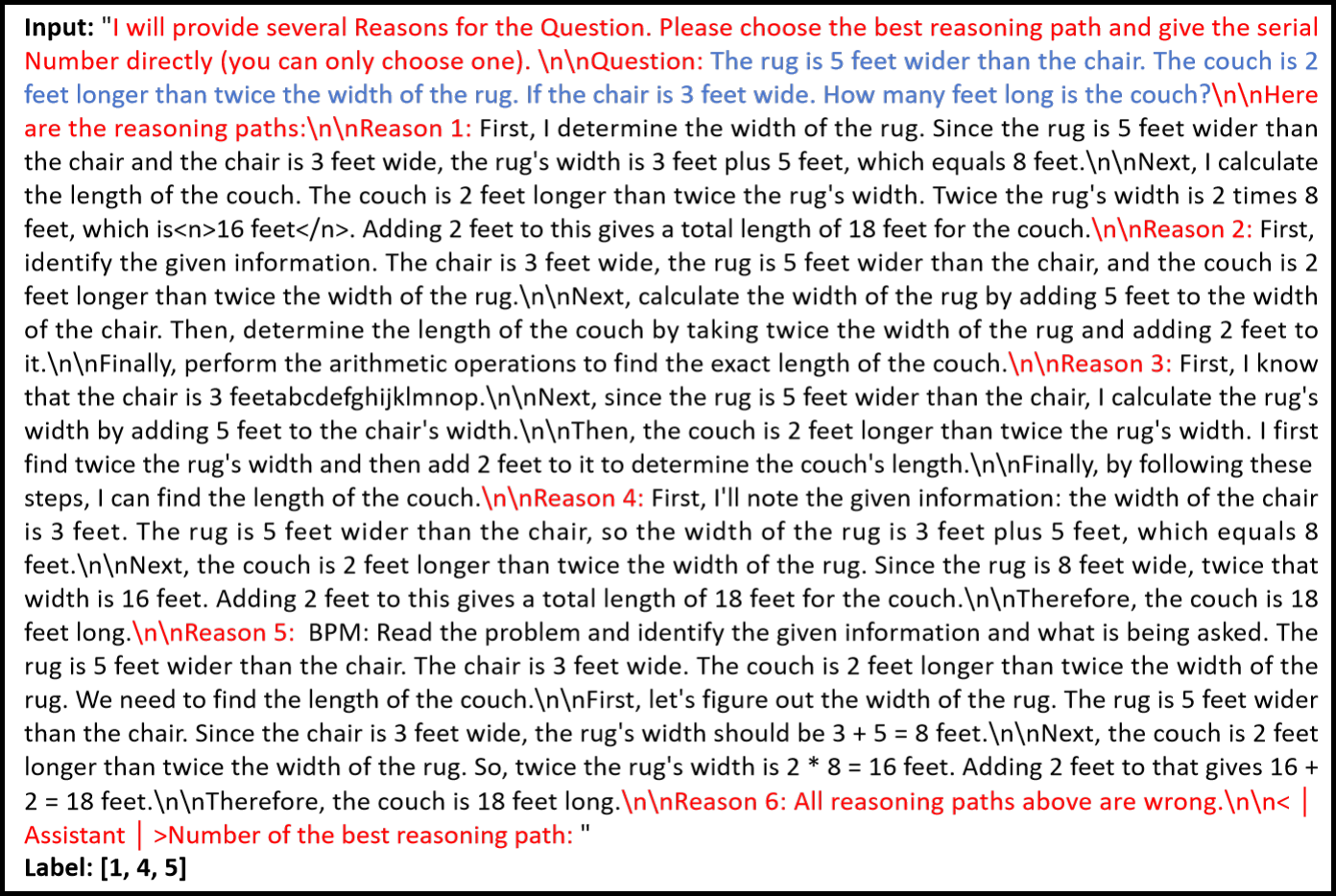}
    \caption{\label{fig:sp1} The selection template is in red and question is in blue.}
\end{figure}

\begin{figure}[h!]
    \centering
    \includegraphics[width=\textwidth]{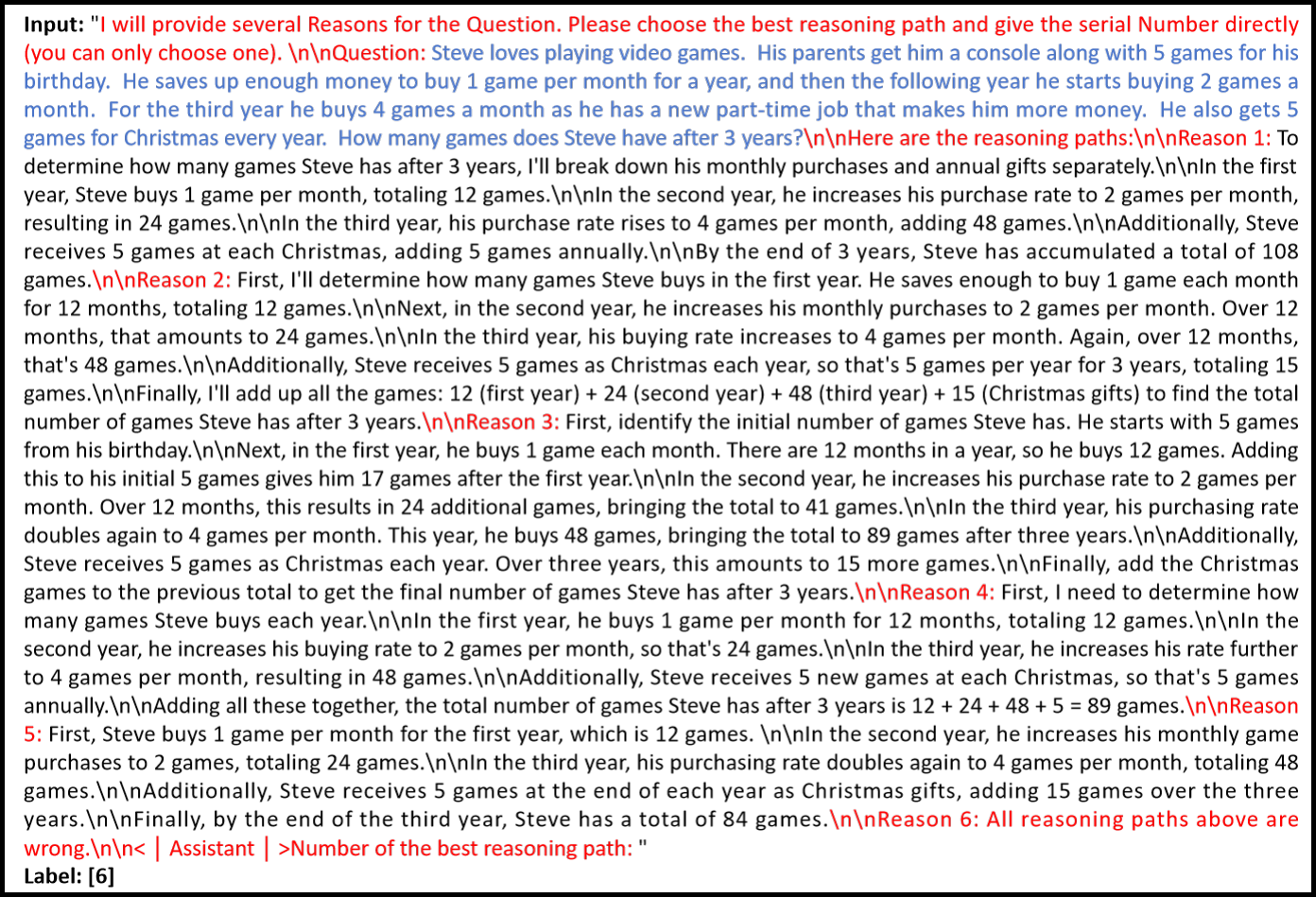}
    \caption{\label{fig:sp2} The selection template is in red and question is in blue.}
\end{figure}

We display two samples of constructed training data for draft selection in Figure \ref{fig:sp1} and Figure \ref{fig:sp2}.

\section{Limitations}
\label{sec:limitation}
\begin{itemize}[left=0pt]
\item SCoT adopts existing pre-trained reasoning models as draft models.
However, for some target models, existing pre-trained models of the same family may not be available.
In addition, the size of the draft model has a significant impact on the efficiency of reasoning. 
For example, in the experiments, the speedup ratio of SCoT on the Llama group is lower than that on the Qwen group. This is largely because the 8B draft model is still a bit large for the 70B target model.
An overly large draft model may limit acceleration performance or even slow down reasoning. Future work can consider how to train a more effective draft model for any target model.
\item SCoT needs to fine-tune the main model and the target model. In this paper, due to resource limitations, we only consider its performance on mathematical reasoning tasks.
We used only limited training data.
Its performance on other types of reasoning tasks may vary, especially for out-of-domain tasks.
\item SCoT may change the original output of the target model, which may have a certain impact on the original security of the pre-trained model. In practical applications, it may lead to harmful output.
If used in real-world scenarios, the security of the inference framework needs to be reconsidered.
\end{itemize}

\newpage
\section*{NeurIPS Paper Checklist}

\begin{enumerate}

\item {\bf Claims}
    \item[] Question: Do the main claims made in the abstract and introduction accurately reflect the paper's contributions and scope?
    \item[] Answer: \answerYes{} 
    \item[] Justification: We propose a method for improving the reasoning efficiency of LLMs. We clearly state our goal and contributions in the abstract and introduction.
    \item[] Guidelines:
    \begin{itemize}
        \item The answer NA means that the abstract and introduction do not include the claims made in the paper.
        \item The abstract and/or introduction should clearly state the claims made, including the contributions made in the paper and important assumptions and limitations. A No or NA answer to this question will not be perceived well by the reviewers. 
        \item The claims made should match theoretical and experimental results, and reflect how much the results can be expected to generalize to other settings. 
        \item It is fine to include aspirational goals as motivation as long as it is clear that these goals are not attained by the paper. 
    \end{itemize}

\item {\bf Limitations}
    \item[] Question: Does the paper discuss the limitations of the work performed by the authors?
    \item[] Answer: \answerYes{} 
    \item[] Justification: We discussed our limitation in Appendix \ref{sec:limitation}.
    \item[] Guidelines:
    \begin{itemize}
        \item The answer NA means that the paper has no limitation while the answer No means that the paper has limitations, but those are not discussed in the paper. 
        \item The authors are encouraged to create a separate "Limitations" section in their paper.
        \item The paper should point out any strong assumptions and how robust the results are to violations of these assumptions (e.g., independence assumptions, noiseless settings, model well-specification, asymptotic approximations only holding locally). The authors should reflect on how these assumptions might be violated in practice and what the implications would be.
        \item The authors should reflect on the scope of the claims made, e.g., if the approach was only tested on a few datasets or with a few runs. In general, empirical results often depend on implicit assumptions, which should be articulated.
        \item The authors should reflect on the factors that influence the performance of the approach. For example, a facial recognition algorithm may perform poorly when image resolution is low or images are taken in low lighting. Or a speech-to-text system might not be used reliably to provide closed captions for online lectures because it fails to handle technical jargon.
        \item The authors should discuss the computational efficiency of the proposed algorithms and how they scale with dataset size.
        \item If applicable, the authors should discuss possible limitations of their approach to address problems of privacy and fairness.
        \item While the authors might fear that complete honesty about limitations might be used by reviewers as grounds for rejection, a worse outcome might be that reviewers discover limitations that aren't acknowledged in the paper. The authors should use their best judgment and recognize that individual actions in favor of transparency play an important role in developing norms that preserve the integrity of the community. Reviewers will be specifically instructed to not penalize honesty concerning limitations.
    \end{itemize}

\item {\bf Theory assumptions and proofs}
    \item[] Question: For each theoretical result, does the paper provide the full set of assumptions and a complete (and correct) proof?
    \item[] Answer: \answerNA{} 
    \item[] Justification: This paper does not include theoretical results.
    \item[] Guidelines:
    \begin{itemize}
        \item The answer NA means that the paper does not include theoretical results. 
        \item All the theorems, formulas, and proofs in the paper should be numbered and cross-referenced.
        \item All assumptions should be clearly stated or referenced in the statement of any theorems.
        \item The proofs can either appear in the main paper or the supplemental material, but if they appear in the supplemental material, the authors are encouraged to provide a short proof sketch to provide intuition. 
        \item Inversely, any informal proof provided in the core of the paper should be complemented by formal proofs provided in appendix or supplemental material.
        \item Theorems and Lemmas that the proof relies upon should be properly referenced. 
    \end{itemize}

    \item {\bf Experimental result reproducibility}
    \item[] Question: Does the paper fully disclose all the information needed to reproduce the main experimental results of the paper to the extent that it affects the main claims and/or conclusions of the paper (regardless of whether the code and data are provided or not)?
    \item[] Answer: \answerYes{} 
    \item[] Justification: We described the architecture and experimental details in Section \ref{sec:scot} and \ref{sec:ex}. We also provide our code in supplementary materials.
    \item[] Guidelines:
    \begin{itemize}
        \item The answer NA means that the paper does not include experiments.
        \item If the paper includes experiments, a No answer to this question will not be perceived well by the reviewers: Making the paper reproducible is important, regardless of whether the code and data are provided or not.
        \item If the contribution is a dataset and/or model, the authors should describe the steps taken to make their results reproducible or verifiable. 
        \item Depending on the contribution, reproducibility can be accomplished in various ways. For example, if the contribution is a novel architecture, describing the architecture fully might suffice, or if the contribution is a specific model and empirical evaluation, it may be necessary to either make it possible for others to replicate the model with the same dataset, or provide access to the model. In general. releasing code and data is often one good way to accomplish this, but reproducibility can also be provided via detailed instructions for how to replicate the results, access to a hosted model (e.g., in the case of a large language model), releasing of a model checkpoint, or other means that are appropriate to the research performed.
        \item While NeurIPS does not require releasing code, the conference does require all submissions to provide some reasonable avenue for reproducibility, which may depend on the nature of the contribution. For example
        \begin{enumerate}
            \item If the contribution is primarily a new algorithm, the paper should make it clear how to reproduce that algorithm.
            \item If the contribution is primarily a new model architecture, the paper should describe the architecture clearly and fully.
            \item If the contribution is a new model (e.g., a large language model), then there should either be a way to access this model for reproducing the results or a way to reproduce the model (e.g., with an open-source dataset or instructions for how to construct the dataset).
            \item We recognize that reproducibility may be tricky in some cases, in which case authors are welcome to describe the particular way they provide for reproducibility. In the case of closed-source models, it may be that access to the model is limited in some way (e.g., to registered users), but it should be possible for other researchers to have some path to reproducing or verifying the results.
        \end{enumerate}
    \end{itemize}

\item {\bf Open access to data and code}
    \item[] Question: Does the paper provide open access to the data and code, with sufficient instructions to faithfully reproduce the main experimental results, as described in supplemental material?
    \item[] Answer: \answerYes{} 
    \item[] Justification: We provide our code in supplementary materials.
    \item[] Guidelines:
    \begin{itemize}
        \item The answer NA means that paper does not include experiments requiring code.
        \item Please see the NeurIPS code and data submission guidelines (\url{https://nips.cc/public/guides/CodeSubmissionPolicy}) for more details.
        \item While we encourage the release of code and data, we understand that this might not be possible, so “No” is an acceptable answer. Papers cannot be rejected simply for not including code, unless this is central to the contribution (e.g., for a new open-source benchmark).
        \item The instructions should contain the exact command and environment needed to run to reproduce the results. See the NeurIPS code and data submission guidelines (\url{https://nips.cc/public/guides/CodeSubmissionPolicy}) for more details.
        \item The authors should provide instructions on data access and preparation, including how to access the raw data, preprocessed data, intermediate data, and generated data, etc.
        \item The authors should provide scripts to reproduce all experimental results for the new proposed method and baselines. If only a subset of experiments are reproducible, they should state which ones are omitted from the script and why.
        \item At submission time, to preserve anonymity, the authors should release anonymized versions (if applicable).
        \item Providing as much information as possible in supplemental material (appended to the paper) is recommended, but including URLs to data and code is permitted.
    \end{itemize}

\item {\bf Experimental setting/details}
    \item[] Question: Does the paper specify all the training and test details (e.g., data splits, hyperparameters, how they were chosen, type of optimizer, etc.) necessary to understand the results?
    \item[] Answer: \answerYes{} 
    \item[] Justification: We provided the training and test details in Section \ref{sec:ex}. The way to chose the core hyperparameter is stated in Section \ref{sec:chain}.
    \item[] Guidelines:
    \begin{itemize}
        \item The answer NA means that the paper does not include experiments.
        \item The experimental setting should be presented in the core of the paper to a level of detail that is necessary to appreciate the results and make sense of them.
        \item The full details can be provided either with the code, in appendix, or as supplemental material.
    \end{itemize}

\item {\bf Experiment statistical significance}
    \item[] Question: Does the paper report error bars suitably and correctly defined or other appropriate information about the statistical significance of the experiments?
    \item[] Answer: \answerYes{} 
    \item[] Justification: The results are accompanied by bars and tables in Section \ref{sec:ex}.
    \item[] Guidelines:
    \begin{itemize}
        \item The answer NA means that the paper does not include experiments.
        \item The authors should answer "Yes" if the results are accompanied by error bars, confidence intervals, or statistical significance tests, at least for the experiments that support the main claims of the paper.
        \item The factors of variability that the error bars are capturing should be clearly stated (for example, train/test split, initialization, random drawing of some parameter, or overall run with given experimental conditions).
        \item The method for calculating the error bars should be explained (closed form formula, call to a library function, bootstrap, etc.)
        \item The assumptions made should be given (e.g., Normally distributed errors).
        \item It should be clear whether the error bar is the standard deviation or the standard error of the mean.
        \item It is OK to report 1-sigma error bars, but one should state it. The authors should preferably report a 2-sigma error bar than state that they have a 96\% CI, if the hypothesis of Normality of errors is not verified.
        \item For asymmetric distributions, the authors should be careful not to show in tables or figures symmetric error bars that would yield results that are out of range (e.g. negative error rates).
        \item If error bars are reported in tables or plots, The authors should explain in the text how they were calculated and reference the corresponding figures or tables in the text.
    \end{itemize}

\item {\bf Experiments compute resources}
    \item[] Question: For each experiment, does the paper provide sufficient information on the computer resources (type of compute workers, memory, time of execution) needed to reproduce the experiments?
    \item[] Answer: \answerYes{} 
    \item[] Justification: We reported the resources used in experiments in Section \ref{sec:ex}.
    \item[] Guidelines:
    \begin{itemize}
        \item The answer NA means that the paper does not include experiments.
        \item The paper should indicate the type of compute workers CPU or GPU, internal cluster, or cloud provider, including relevant memory and storage.
        \item The paper should provide the amount of compute required for each of the individual experimental runs as well as estimate the total compute. 
        \item The paper should disclose whether the full research project required more compute than the experiments reported in the paper (e.g., preliminary or failed experiments that didn't make it into the paper). 
    \end{itemize}
    
\item {\bf Code of ethics}
    \item[] Question: Does the research conducted in the paper conform, in every respect, with the NeurIPS Code of Ethics \url{https://neurips.cc/public/EthicsGuidelines}?
    \item[] Answer: \answerYes{} 
    \item[] Justification: We have reviewed the NeurIPS Code of Ethics.
    \item[] Guidelines:
    \begin{itemize}
        \item The answer NA means that the authors have not reviewed the NeurIPS Code of Ethics.
        \item If the authors answer No, they should explain the special circumstances that require a deviation from the Code of Ethics.
        \item The authors should make sure to preserve anonymity (e.g., if there is a special consideration due to laws or regulations in their jurisdiction).
    \end{itemize}

\item {\bf Broader impacts}
    \item[] Question: Does the paper discuss both potential positive societal impacts and negative societal impacts of the work performed?
    \item[] Answer: \answerYes{} 
    \item[] Justification: We discuss the societal impacts in Appendix \ref{sec:limitation}.
    \item[] Guidelines:
    \begin{itemize}
        \item The answer NA means that there is no societal impact of the work performed.
        \item If the authors answer NA or No, they should explain why their work has no societal impact or why the paper does not address societal impact.
        \item Examples of negative societal impacts include potential malicious or unintended uses (e.g., disinformation, generating fake profiles, surveillance), fairness considerations (e.g., deployment of technologies that could make decisions that unfairly impact specific groups), privacy considerations, and security considerations.
        \item The conference expects that many papers will be foundational research and not tied to particular applications, let alone deployments. However, if there is a direct path to any negative applications, the authors should point it out. For example, it is legitimate to point out that an improvement in the quality of generative models could be used to generate deepfakes for disinformation. On the other hand, it is not needed to point out that a generic algorithm for optimizing neural networks could enable people to train models that generate Deepfakes faster.
        \item The authors should consider possible harms that could arise when the technology is being used as intended and functioning correctly, harms that could arise when the technology is being used as intended but gives incorrect results, and harms following from (intentional or unintentional) misuse of the technology.
        \item If there are negative societal impacts, the authors could also discuss possible mitigation strategies (e.g., gated release of models, providing defenses in addition to attacks, mechanisms for monitoring misuse, mechanisms to monitor how a system learns from feedback over time, improving the efficiency and accessibility of ML).
    \end{itemize}
    
\item {\bf Safeguards}
    \item[] Question: Does the paper describe safeguards that have been put in place for responsible release of data or models that have a high risk for misuse (e.g., pretrained language models, image generators, or scraped datasets)?
    \item[] Answer: \answerYes{} 
    \item[] Justification: We discuss the safeguards in Appendix \ref{sec:limitation}.
    \item[] Guidelines:
    \begin{itemize}
        \item The answer NA means that the paper poses no such risks.
        \item Released models that have a high risk for misuse or dual-use should be released with necessary safeguards to allow for controlled use of the model, for example by requiring that users adhere to usage guidelines or restrictions to access the model or implementing safety filters. 
        \item Datasets that have been scraped from the Internet could pose safety risks. The authors should describe how they avoided releasing unsafe images.
        \item We recognize that providing effective safeguards is challenging, and many papers do not require this, but we encourage authors to take this into account and make a best faith effort.
    \end{itemize}

\item {\bf Licenses for existing assets}
    \item[] Question: Are the creators or original owners of assets (e.g., code, data, models), used in the paper, properly credited and are the license and terms of use explicitly mentioned and properly respected?
    \item[] Answer: \answerYes{} 
    \item[] Justification: We cited the original paper of models, datasets and codes used in this paper. We respect their license and terms.
    \item[] Guidelines:
    \begin{itemize}
        \item The answer NA means that the paper does not use existing assets.
        \item The authors should cite the original paper that produced the code package or dataset.
        \item The authors should state which version of the asset is used and, if possible, include a URL.
        \item The name of the license (e.g., CC-BY 4.0) should be included for each asset.
        \item For scraped data from a particular source (e.g., website), the copyright and terms of service of that source should be provided.
        \item If assets are released, the license, copyright information, and terms of use in the package should be provided. For popular datasets, \url{paperswithcode.com/datasets} has curated licenses for some datasets. Their licensing guide can help determine the license of a dataset.
        \item For existing datasets that are re-packaged, both the original license and the license of the derived asset (if it has changed) should be provided.
        \item If this information is not available online, the authors are encouraged to reach out to the asset's creators.
    \end{itemize}

\item {\bf New assets}
    \item[] Question: Are new assets introduced in the paper well documented and is the documentation provided alongside the assets?
    \item[] Answer: \answerYes{} 
    \item[] Justification: We provide documentation alongside the codes in supplementary materials.
    \item[] Guidelines: 
    \begin{itemize}
        \item The answer NA means that the paper does not release new assets.
        \item Researchers should communicate the details of the dataset/code/model as part of their submissions via structured templates. This includes details about training, license, limitations, etc. 
        \item The paper should discuss whether and how consent was obtained from people whose asset is used.
        \item At submission time, remember to anonymize your assets (if applicable). You can either create an anonymized URL or include an anonymized zip file.
    \end{itemize}

\item {\bf Crowdsourcing and research with human subjects}
    \item[] Question: For crowdsourcing experiments and research with human subjects, does the paper include the full text of instructions given to participants and screenshots, if applicable, as well as details about compensation (if any)? 
    \item[] Answer: \answerNA{} 
    \item[] Justification: This paper does not involve crowdsourcing nor research with human subjects.
    \item[] Guidelines:
    \begin{itemize}
        \item The answer NA means that the paper does not involve crowdsourcing nor research with human subjects.
        \item Including this information in the supplemental material is fine, but if the main contribution of the paper involves human subjects, then as much detail as possible should be included in the main paper. 
        \item According to the NeurIPS Code of Ethics, workers involved in data collection, curation, or other labor should be paid at least the minimum wage in the country of the data collector. 
    \end{itemize}

\item {\bf Institutional review board (IRB) approvals or equivalent for research with human subjects}
    \item[] Question: Does the paper describe potential risks incurred by study participants, whether such risks were disclosed to the subjects, and whether Institutional Review Board (IRB) approvals (or an equivalent approval/review based on the requirements of your country or institution) were obtained?
    \item[] Answer: \answerNA{} 
    \item[] Justification: This paper does not involve crowdsourcing nor research with human subjects.
    \item[] Guidelines:
    \begin{itemize}
        \item The answer NA means that the paper does not involve crowdsourcing nor research with human subjects.
        \item Depending on the country in which research is conducted, IRB approval (or equivalent) may be required for any human subjects research. If you obtained IRB approval, you should clearly state this in the paper. 
        \item We recognize that the procedures for this may vary significantly between institutions and locations, and we expect authors to adhere to the NeurIPS Code of Ethics and the guidelines for their institution. 
        \item For initial submissions, do not include any information that would break anonymity (if applicable), such as the institution conducting the review.
    \end{itemize}

\item {\bf Declaration of LLM usage}
    \item[] Question: Does the paper describe the usage of LLMs if it is an important, original, or non-standard component of the core methods in this research? Note that if the LLM is used only for writing, editing, or formatting purposes and does not impact the core methodology, scientific rigorousness, or originality of the research, declaration is not required.
    \item[] Answer: \answerNA{} 
    \item[] Justification: The core method development in this research does not involve LLMs as any important, original, or non-standard components.
    \item[] Guidelines:
    \begin{itemize}
        \item The answer NA means that the core method development in this research does not involve LLMs as any important, original, or non-standard components.
        \item Please refer to our LLM policy (\url{https://neurips.cc/Conferences/2025/LLM}) for what should or should not be described.
    \end{itemize}

\end{enumerate}

\end{document}